\documentclass[11pt,a4paper]{article}
\usepackage[hyperref]{emnlp2019}
\usepackage{times}
\usepackage{latexsym}
\usepackage{url}
\usepackage{graphicx}
\usepackage{dirtytalk}
\usepackage{booktabs} 
\usepackage{caption} 
\usepackage{amsmath}
\usepackage{amsfonts}
\usepackage{algorithm}
\usepackage{algpseudocode}
\usepackage{array,multirow}
\usepackage{xcolor}

\DeclareMathOperator*{\argmax}{arg\,max} 

\usepackage{amsthm}
\newtheorem{theorem}{Theorem}
 
\newtheorem{definition}[theorem]{Definition}
\usepackage[colorinlistoftodos]{todonotes}
\usepackage{enumitem}
\newlist{Properties}{enumerate}{2}
\setlist[Properties]{label=Property \arabic*.,itemindent=*}

\usepackage{booktabs}       
\usepackage{nicefrac}       

\aclfinalcopy 

\title{Error-Correcting Neural Sequence Prediction}

\author{James O' Neill and Danushka Bollegala\\
  Department of Computer Science, University of Liverpool\\
  Liverpool, L69 3BX \\
  England \\
  {\tt \{james.o-neill, danushka.bollegala\}@liverpool.ac.uk} \\}

\date{}

\begin{document}
\maketitle

\begin{abstract}
We propose a novel neural sequence prediction method based on \textit{error-correcting output codes} that avoids exact softmax normalization and allows for a tradeoff between speed and performance. Instead of minimizing measures between the predicted probability distribution and true distribution, we use error-correcting codes to represent both predictions and outputs. 
Secondly, we propose multiple ways to improve accuracy and convergence rates by maximizing the separability between codes that correspond to classes proportional to word embedding similarities. 

Lastly, we introduce our main contribution called \textit{Latent Variable Mixture Sampling}, a technique that is used to mitigate exposure bias, which can be integrated into training latent variable-based neural sequence predictors such as ECOC.
This involves mixing the latent codes of past predictions and past targets in one of two ways: (1) according to a predefined sampling schedule or (2) a differentiable sampling procedure whereby the mixing probability is learned throughout training by replacing the greedy argmax operation with a smooth approximation. ECOC-NSP leads to consistent improvements on language modelling datasets and the proposed Latent Variable mixture sampling methods are found to perform well for text generation tasks such as image captioning.
\end{abstract}

\section{Introduction}
Sequence modelling (SM) is a fundamental task in natural language which requires a parametric model to generate tokens given past tokens. SM underlies many types of structured prediction tasks in natural language, such as Language Modelling (LM), Sequence Tagging (e.g Named Entity Recognition, Constituency/Dependency Parsing) and Text Generation (e.g Image Captioning and Machine  Answering~\citep{sutskever2014sequence}).  
The goal is to learn a joint probability distribution for a sequence of length $T$ containing words from a vocabulary $\mathcal{V}$. This distribution can be decomposed into the conditional distributions of current tokens given past tokens using the chain rule shown in Equation \ref{eq:cond_dist}. In Neural Sequence Prediction (NSP), a Recurrent Neural Network (RNN) $f_{\theta}(\cdot)$ parameterized by $\theta$ is used to encode the information at each timestep $t$ into a hidden state vector $h^{l}_t$ which is followed by a decoder $z^{l}_t = h^{l}_t W^{l} + b^{l}$ and a normalization function $\phi(z_{t}^{l})$ which forms a probability distribution $\hat{p}_{\theta}(y_t|x_t, h_{t-1})$, $\; \forall t \in [0,1,..T]$. 

\begin{equation}\label{eq:cond_dist}
    P(w_1, ..., w_T) = \prod^{T}_{t=1} P(w_t| w_{t−1},..., w_1)
\end{equation}

However, training can be slow when $|\mathcal{V}|$ is large while also leaving a large memory footprint for the respective input embedding matrices. Conversely, in cases where the decoder is limited by an information bottleneck, the opposite is required where more degrees of freedom are necessary to alleviate information loss in the decoder bottleneck. Both scenarios correspond to a trade-off between computation complexity and out-of-sample performance. Hence, we require that a newly proposed model has the property that the decoder can be easily configured to deal with this trade-off in a principled way.
Lastly, standard supervised learning (self-supervised for sequence prediction) assumes inputs are i.i.d. However, in sequence prediction, the model has to rely on its own predictions at test time, instead of past targets that are used as input at training time. This difference is known as \textit{exposure bias} and can lead to errors compounding along a generated sequence. This approach to sequence prediction is also known as \textit{teacher forcing} where the teacher provides all targets at training time. We also require that exposure bias is addressed while dealing with the aforementioned challenges related to computation and performance trade-offs in the decoder.
\newline
Hence, we propose an error-correcting output code (ECOC) based NSP (ECOC-NSP) that address this desideratum. We show that when given sufficient error codes ($ |\mathcal{V}| \gg |c| \gg \log_2(|\mathcal{V}|) $) while the codeword dimensionality $|c| < |\mathcal{V}|$, accuracy can be maintained compared to using the full softmax.
Lastly, we show that this latent variable-based NSP approach can be extended to mitigate the aforementioned problem of compounding errors by using \textit{Latent Variable Mixture Sampling} (LVMS). LVMS in an ECOC-NSP model also outperforms an equivalent Hierarchical Softmax-based NSP that uses Scheduled Sampling~\citep{bengio2015scheduled} and other closely related baselines. This is the first report of  mitigating compounding errors when approximating the posterior in recurrent neural networks (RNNs).

\paragraph{Contributions} Our main contributions are summarized as the following:

\begin{enumerate}
    \itemsep0em 
    \item An error-correcting output coded neural language model that requires less parameters than its softmax-based sequence modelling counterpart given sufficient separability between classes via error-checks.
    \item An embedding cosine similarity rank ordered codebook that leads to well-separated codewords, where the number of error-correcting codes assigned to a token is proportional to the cosine similarity between the tokens corresponding pretrained word embedding and the most frequent tokens word embedding.
    \item A \textit{Latent-Mixture Sampling} method to mitigate exposure bias in Latent Variable models. This is then extended to \textit{Differentiable Latent Variable Mixture Sampling} that uses the Gumbel-Softmax so that discrete categorical variables can be backpropogated through. This performs comparably to other sampling-based approaches.
    \item Novel baselines such as Scheduled Sampling Hierarchical Softmax (SS-HS) and Scheduled Sampling Adaptive Softmax (SS-AS), are introduced in the evaluation of our proposed ECOC method. This applies SS to closely related softmax approximations. 
\end{enumerate}

\section{Background}
\subsection{Error-Correcting Codes}
Error-Correcting Codes~\citep{hamming1950error} originate from seminal work in solid-state electronics around the time of the first digital computer. Later,  binary codes were introduced in the context of artificial intelligence via the NETtalk system~\citep{sejnowski1987parallel}, where each class index is represented by a binary code $C$ and a predicted code as $\hat{C}$ from some parametric model $f_{\theta}(\cdot)$. When $|\mathcal{V}| \neq 2^{n}$, the remaining codes are used as error-correction bits $k = |\mathcal{V}| - 2^{n}$. This tolerance can be used to account for the information loss due to the sample size by measuring the distance (e.g Hamming) between the predicted codeword and the true codeword with $d$ error-correction bits. If the minimum distance between codewords is $d$ then at least $(d-1)/2$ bits can be corrected for and hence, if the Hamming distance $d \leq (d-1)/2$ we will still retrieve the correct codeword. In contrast to using one bit per $k$ classes in standard multi-class classification, error-correction cannot be achieved. Both error-correction and class bits make up the \textit{codebook} $\mathcal{C}$. 

\paragraph{\textit{Why Latent Codes for Neural Sequence Prediction?}} Targets are represented as 1-hot vectors (i.e. kronecker delta) in standard training of neural sequence predictors, treated as a 1-vs-rest multi-class classification. This approach can be considered a special case of ECOC classification where the codebook $\mathcal{C}$ with $n$ classes is represented by an identity $\mathbb{I}^{n \times n}$. ECOC classification is well suited over explicitly using observed variables when the output space is structured. Flat-classification (1-vs-rest) ignores the dependencies in the outputs, in comparison to using latent codes that share some common latent variables between associative words. For example, in the former case, if we train a model that only observes the word \say{silver} in a sequence \say{...silver car..} and then at test-time observes \say{silver-back}, because there is high association between \say{silver} and \say{car}, the model is more likely to predict \say{car} instead of \say{gorilla}. ECOC is less prone to such mistakes because although a/some bit/s may be different between the latent codes for \say{car} and \say{gorilla}, the potential misclassifications can be re-corrected with the error-correcting bits. In other words, latent coding can reduce the variance of each individual classifier and has some tolerance to mistakes induced by sparse transitions, proportional to the number of error-checks used.

\subsection{Methods for Softmax Approximation}

~\newcite{goodman2001bit,morin2005hierarchical} proposed a \textbf{Hierarchical Softmax} (HS) that outputs short codes representing marginals of the conditional distribution, where the product of these marginals along a path in the tree approximate the conditional distribution. This speeds up training proportional to the traversed binary tree path lengths, where intermediate nodes assign relative probabilities of child nodes.
Defining a good tree structure improves performance since semantically similar words have a shorter path and therefore similar representations are learned for similar words. HS and ECOC are similar insofar as they both can be interpreted as approximating the posterior as a product of marginal probabilities using short codes. However, ECOC tolerates errors in code prediction. This becomes more important as the code length (i.e the tree depth in HS) grows since the likelihood of mistakes becomes higher. Hence, ECOC can be considered a flexible tradeoff between the strictness of HS and the full softmax. 
The \textbf{Differentiated Softmax} (DS) uses a sparse linear block of weights for the decoder where a set of partitions are made according to the unigram distribution, where the number of weights are assigned proportional to the term frequency. This is intuitive since rare words require less degrees of freedom to account for the little amount of contexts in which they appear, compared to common words. The optimal spacing of error-checking bits between codewords corresponding to tokens is similar to - varying branching sizes for different areas of the tree depending on the unigram frequency or allocating weights proportional to the frequency akin to Differentiable Softmax~\citep{chen2015strategies}. We also consider achieving the spacing via term frequency, but also use our proposed rank ordered word embedding cosine similarities as mentioned in Equation \ref{sec:cc}.

The \textbf{Adaptive Softmax} (AS) ~\citet{grave2016efficient} provide an approximate hierarchical model that directly accounts for the computation time of matrix multiplications. AS results in 2x-10x speedups when compared to the standard softmax, dependent on the size of the corpus and vocabulary. They find on sufficiently large corpora (Text8, Europarl and 1-Billion datasets), accuracy is maintained while reducing the computation time.

\subsection{Recent Applications of Latent Codes}\label{sec:rr}
~\citet{shu2017compressing} recently used compositional codes for word embeddings to cut down on memory requirements in mobile devices. Instead of using binary coding, they achieve word embedding compression using multi-codebook quantization. Each bit $c \in C$ comprises of a discrete code (0-9) and therefore at minimum $\log_{10}(k)$ bits are required. They also propose to use the Gumbel-Softmax trick but for the purposes of learning the discrete codes. The performance was maintaned for sentiment analysis and machine translation with 94\% and 98\% respective compression rates.
~\citet{shi2018structured} propose a product quantizatioon structured embedding that reduces memory by 10-20 times the number of parameters, while maintaining performance. This involves slicing the embedding tensor into groups which are then quantized independently. 
~\cite{oda2017neural} also consider binary code for neural machine translation. However, their approach still required binary codes to be used in conjunction with the softmax and showed a performance degradation when only using binary codes. Here, we show that, when given enough bits, the model is competitive against the full softmax, and in some cases outperforming. Moreover, we introduce novel ways to mitigate exposure bias in these models, and Latent Variable based models alike.

\section{Methodology}
\subsection{Codebook Construction}\label{sec:cc}

A challenging aspect of assigning codewords is ordering the codes so that even if incorrect predictions are made, that the codeword is at least semantically closer to that of the codewords that are less related, while ensuring good separation between codes. Additionally, we have to consider the amount of error-checking bits to use. In theory, $\log_2(k)/k$ is sufficient to account for all $k$ classes. However, this alone can lead a degradation in performance. Hence, we also consider a large amount of error-checking bits. In this case, the error-checking bits can account for more mistakes given by other classes, which may be correlated. In contrast, using probability distributions naturally accounts for these correlations, as the mass needs to shift relative to the activation of each output.  
This is particularly important for language modelling and text generation because of the high-dimensionality of the output. The most naive way to create the codebook is to assign binary codes to each word in random order. However, it is preferable to assign similar codes to $w \in \mathcal{V}$ that are semantically similar while maximizing the Hamming distance between codes where leftover error codes separate class codes.
    
\subsubsection{Codebook Arrangement}
A fundamental challenge in creating the codebook $\mathcal{C}$ is in how error-codes are distributed between codes that maximize the separability between codewords that are more likely to be interchangeably and incorrectly predicted. This is related to the second challenge of choosing the dimensionality of $C$. The latter is dependent on the size of the corpus, and in some cases might only require $|\log_2(\mathcal{V})| \leq d \leq |\mathcal{V}|$ bits to represent all classes with leftover error-checking bits. These two decisions correspond to a tradeoff between computational complexity and accuracy of our neural language model, akin to tree expressitivity in the Hierarhcial Softmax to using the Full Softmax. Below we describe a semantically motivated method to achieve well-separated codewords, followed by a guide on how to choose codebook dimensionality $d_\mathcal{C}$.

\paragraph{Embedding Simlarity-Based Codebooks}\label{sec:esbc}
Previous work on ECOC has focused on theories as to why randomly generated codes lead to good row and column separation~\citep{berger1999error}. However, this assumes that class labels are conditionally independent and therefore it does not apply well for sequence modelling where the output space is loosely structured. To address this, we propose to reorder $C \in \mathcal{C}$ such that Hamming distance between any two codewords is proportional to the embedding similarity.
Moreover, separating codewords by semantic similarity can be achieved by placing the amount of error-checking bits proportional to rank ordered similarity for a chosen query word embedding.
A codebook ordered by pretrained word embedding similarities for $w_{*}$ is denoted as $\mathcal{C}_{\Lambda_{w_{*}}}$. The similarity scores between embeddings is given as $\mathcal{F}(\mathcal{M}_{*}, M_{i}) \; \forall i$ is used reorder $M \to \mathcal{M}^{'}$. In our experiments we use pretrained GoogleNews $\texttt{skipgram}$ embeddings\footnote{available here: \url{https://code.google.com/archive/p/word2vec/}}. 

Good separation is achieved when codes are separated proportional to the cosine similarity between embeddings of the most frequent word $w_{*} \in \mathcal{V}$ and the remaining words $w'$. Therefore, words with high similarity have corresponding codes that are closer in Hamming distance $\text{H}(\cdot,\cdot)$ in $\mathcal{C}$. 
This ensures that even when codes are correlated, that incorrect latent predictions are at least more likely to correspond to semantically related words. We are not guaranteed that codes close in Hamming distance are closer in a semantic sense in the random case. Therefore, we can instead consider computing $\mathcal{M}^{'}$ using ordered similarities of word embedding 
similarities where the function $\mathcal{F}(\cdot, \cdot)$ computes the cosine similarity for any two words. 
If we apply the same combinatorial analysis to codebook separation when the codes are assigned proportional to semantic similarity, we can define the good separation as the following: 

\begin{definition}
Given $k$ redundant codewords $C_k$, we require an assignment that leads to a strongly separated $\mathcal{C}$. Let $\delta(\cdot, \cdot)$ denote a function that assigns $C_{i}^k$ error-checking codewords assigned to the $i^{th}$ class codeword and $\delta(C_{*}, C_{i}) \propto \mathcal{F}(\mathcal{M}_{*}, \mathcal{M}_{i}) \; \forall k$.
\end{definition}
In practice $\delta(\cdot, \cdot)$ normalizes the resulting embedding similarities $S = \mathcal{F}(M_{*}, M)$ using a normalization function cumsum $\big(\phi(S)\times |\mathcal{C}|\big)$ to assign the intervals between adjacent codeword spans. This assigns greater distance between words that are more similar to $w*$, and less error-checking codewords to rare words that have distant neighbouring words in the embedding space.

\subsection{Latent Variable Mixture Sampling}\label{sec:LVMS}
To mitigate exposure bias for latent variable-based sequence modelling we propose a sampling strategy that interpolates between predicted and target codewords. We refer to this as Latent Variable Mixture Sampling (LVMS) and its application to ECOC as \textit{Codeword Mixture Sampling} (CMS). 

\paragraph{Curriculum-Based Latent Variable Mixture Sampling}

\begin{figure}
\centering
    \includegraphics[scale=.5]{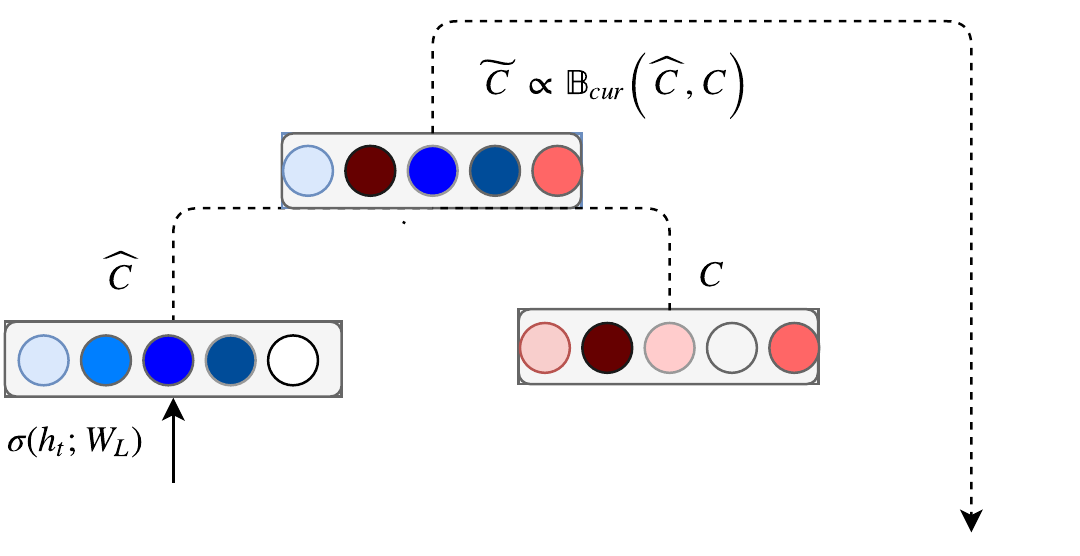}
    \caption{Curriculum Mixture Sampling}\label{fig:cms}
\end{figure}

In Curriculum-Based Latent Variable Mixture Sampling (CLVMS), the mixture probability is $p_{c} = 0 \; \forall c \in C$ at epoch $\epsilon = 0$ and throughout training the probability monotonically increases $p_{c} = \delta_c \; \forall c \in C$ where $\delta_c$ is the threshold for the $c$ th bit after $\epsilon$ epochs.
A Bernoulli sample $\tilde{C} = \mathbb{B}(\hat{C}_{c}, C_{c}) \; \forall c \in [0, C]$ is carried out for $t \in T$ in each minibatch. The probabilities per dimension $p_c$ are independent of keeping a prediction $\hat{y}_{t-1, c}$ instead of the $c$ th bit in the target codeword  $y_{(t-1,c)}$ at timestep $t$-$1$.
The reason for having individual mixture probabilities per bits is because when we consider a default order in $\mathcal{C}$, this results in tokens being assigned codewords ranked by frequency. Therefore, the leftmost bit predictions are more significant than bit errors near the beginning (e.g $2^{0}=1$ only 1 bit difference). We report results when using a sigmoidal schedule as shown in Equation \ref{eq:cur_exp} where $\tau_{max}$ represents the temperature at the last epoch and $\delta$ is a scaling factor controlling the slope.

\begin{equation}\label{eq:cur_exp}
    [\hat{y}_{t-1},y_{t-1}] \sim \frac{\tau_{max}}{1+\exp(-\epsilon/\delta)}, \; \forall \epsilon \in [-\text{N}/2,\text{N}/2]
\end{equation}

This is different to scheduled sampling since we are not just choosing between the ground truth and prediction, instead we are mixing factors of the predicted factored distribution and the target factored distribution that represents the posterior, as illustrated in \autoref{fig:cms}. Here, the color strength illustrates the activation between $[0, 1]$. 
\autoref{fig:cms} also demonstrates how CMS is used in a HS without the additional error checks, in which case each activation corresponding to a node in the tree.

\paragraph{Latent Soft-Mixture Sampling}
In standard CMS, we pass token index $w_t$ which is converted to an input embedding $e_{w}$ based on the most probable bits predictions at the last time step $\argmax_{\theta} p(y_{t-1}|x_{t-1}; \theta)$. We can instead replace the argmax operator with a soft argmax that uses a weighted average of embeddings $e \in E$ where weights are assigned from the previous predicted output via the softmax normalization $\phi(x_{t-1}, \tau)$, where $\tau$ controls the kurtosis of the probability distribution ($\tau \to 0$ tends to argmax) in Equation \ref{eq:soft_argmax}.

\begin{equation}\label{eq:soft_argmax}
     x_t = \sum_{w \in \mathcal{V}} e_{w} \Big( \frac{\exp(h^{T}_{w}\theta/\tau)}{\sum_{w \in \mathcal{V}} \exp(h^{T}_{w}\theta/\tau)}\Big)  
\end{equation}

In the ECOC-NSP, we consider binary codewords and therefore choose the top $k$ least probable bits to flip according to the curriculum schedule. Hence, this results in $k$ codewords where each $\hat{C}$ has at least hamming distance $\text{H}(\hat{C}, C)=1$ $(2^{0})$. Concretely, this is a soft interpolation between past targets and a weighted sum of the $k$ most probable codewords $\hat{C}_K = \argmax_{k}\big(\sigma(h^{T}_{w} W)\big)$ such that $x_t = \mathbb{B}_{K}\Big(C, \sum_k^{K} \phi(\hat{C}_{k})\Big) 
$ where $B_K$ samples one or the other for each $k$th dimension of $C$.

\subsection{Differentiable Latent Variable Sampling}

The previous curriculum strategies disregard where the errors originate from. Instead, they interpolate between model predictions of latent variables $\Hat{Y}$ and targets $Y$ in a way that does not distinguish between cascading errors and localized errors. This means that it only recorrects errors after they are made instead of directly correcting for the origin of the errors. 
~\citet{maddison2016concrete} showed that such operations can be approximated by using a continuous relaxation using the reparameterization trick, also known as the \textit{Concrete Distribution}. By applying such relaxation it allows us to sample from the distribution across codes while allowing for a fully differentiable objective, similar to recent work~\citep{goyal2017differentiable}. 
We extend this to mixture sampling by replacing the argmax operation with the Concrete distribution to allow the gradients to be adjusted at points where prior predictions changed value throughout training. This not only identifies at which time-step the error occurs, but what latent variables (i.e. output codes) had the most influence in generating the error. This is partially motivated by the finding that in the latent variable formulation of simple logistic regression models, the latent variable errors form a Gumbel distribution. Hence, we sample latent codes inversely proportional to the errors from a Gumbel distribution.

\textbf{Gumbel-Softmax}
Similarly, instead of passing the most likely predicted word $\hat{y}^{w*}_{ t-1}$, we can instead sample from $\hat{y}_{t-1} \sim \phi(h_{t-1}, w)$ and then pass this index as $\hat{x}_t$. This is an alternative to always acting greedily and allow the model to seek other likely actions. However, to compute derivatives through samples from the softmax, we need avoid discontinuities, such as the argmax operation. The Gumbel-Softmax~\citep{maddison2016concrete,jang2016categorical} allows us to sample and differentiate through the softmax by providing a continuous relaxation that results in probabilities instead of a step function (i.e. argmax). As shown in \autoref{eq:concrete_dist_tree}, for each componentwise Gumbel noise $k \in [1..,n]$ for latent variable given by $h^{T}\theta$, we find $k$ that maximizes $\log \alpha_k - \log(-\log U_k)$  and then set $D_k=1$ and $D{\neg k} = 0$,  where $U_k \sim \text{Uniform}(0, 1)$ and $\alpha_k$ is drawn from a discrete distribution $D \sim \text{Discrete}(\alpha)$. 

\begin{gather}\label{eq:concrete_dist_tree}
\hat{p}(y_t|x_t;\theta) = \frac{\exp((\log \alpha_k + G_k)/\tau)}{\sum_{i=1}^{n} \exp((\log \alpha_i + G_i)/\tau)}
\end{gather}

For ECOC, we instead consider Bernoulli random variables which for the Concrete distribution can be expressed by means of two arbitrary Gumbel distributions $G_{1}$ and $G_{2}$. The difference between $G_1$ and $G_2$ follows a Logistic distribution and so $G_1 - G_2 \sim \text{Logistic}$ and is sampled as $G_1 - G_2 \equiv \log U - \log(1 - U)$. Hence, if $\alpha = \alpha_1/\alpha_2$, then $P(D_{1} = 1) = P(G_{1} + \log \alpha_{1} > G_{2} + \log \alpha_{2}) = P(\log U − \log(1 - U) + \log \alpha > 0)$. For a step function $\mathcal{H}$, $D_1 \equiv \mathcal{H}(\log \alpha + \log U - \log(1 - U))$, corresponding to the Gumbel Max-Trick~\citep{jang2016categorical}.
\newline
Sampling a Binary Concrete random variable involves sampling $Z$, sample $L \sim \text{Logistic}$ and set $Z$ as shown in \autoref{eq:bin_conc_samp}, where $\alpha, \tau \in (0, \infty)$ and $Z \in (0, 1)$. This Binary Concrete distribution is henceforth denoted as $\text{BinConcrete}(\cdot,\cdot)$ with location $\alpha$ and temperature $\tau$. In the forward pass the probability $Z$ is used to compute the approximate posterior, unlike the one-hot vectors used in straight-through estimation. 

\begin{equation}\label{eq:bin_conc_samp}
Z \equiv \frac{1}{1 + \exp\big(−(\log \alpha + L)/\tau\big)}
\end{equation}

This is used for ECOC and other latent variable-based models, such as Hierarchical Sampling, to propogate through past decisions and make corrective updates that backpropogate to where errors originated from along the sequence. Hence, we also carry out experiments with $\text{BinConcrete}$ (Equation \ref{eq:bin_conc_samp}) and Gumbel-Softmax( Equation \ref{eq:concrete_dist_tree}) for HS and ECOC respectively. The temperature $\tau$ can be kept static, annealed according to a schedule or learned during training, in the latter case this is equivalent to entropy regularization~\citep{grandvalet2005semi} that controls the kurtosis of the distribution. In this work, we consider using an annealed $\tau$, similar to Equation \ref{eq:cur_exp} where $\tau \to 2.5$ and starts with $\tau = 0.01$. This is done to allow the model to avoid large gradient variance in updates early on. In the context of using the Gumbel-Softmax in LVMS, this allows the model to become robust to non-greedy actions gradually throughout training, we would expect such exploration to improve generalization proportional to the vocabulary size.

\section{Experimental Setup}

We carry out experiments for a 2-hidden layer Long-Short Term Memory (LSTM) model with embedding size $|e|=400$, Backpropogation Through Time (BPTT) length $35$ and variational dropout~\cite{gal2016theoretically} with rate $p_d$=0.2 for input, hidden and output layers. The ECOC-NSP model is trained using Binary Cross Entropy loss as shown in \autoref{eq:bce}, where $k$ is a group error-checking codewords corresponding to a codeword $C$. The gradients can then expressed as $\frac{\delta\mathcal{L}}{\delta \theta} = (y−\sigma(h^{T}\theta))\cdot h^{T}$.

\begin{equation}\label{eq:bce}
\begin{gathered}
\mathcal{L}_{\theta}= \max_{k}\prod_{c}^{C}\big[ y_c\log\big(\sigma_c(h^{T}\theta\big)  + \\ (1-y_c)\log\big(1-\sigma_c(h^{T}\theta)\big)\big]  
\end{gathered}
\end{equation}

\textbf{Baselines for ECOC-NSP}
The first set of experiments include comparisons against the most related baselines, which include Sample-Softmax~\citep{bengio2003quick,bengio2008adaptive}, Hierarchical Softmax (HS), AS~\citep{grave2016efficient}, and NCE~\citep{mnih2012fast}. For HS, we use a 2-hidden layer tree with a branching factor (number of classes) of $\sqrt{|\mathcal{V}|}$ by default. For AS, we split the output into 4 groups via the unigram distribution (in percentages of total words 5\%-15\%-30\%-100\%). For NCE, we set the noise ratio to be 0.1 for PTB and 0.2 for WikiText-2 and WikiText-103. Training is carried out until near convergence ($\epsilon \approx 40$), the randomly initialized HS and Sampled Softmax of which take longer ($\epsilon \in$ [55-80]).
Table \ref{tab:ecoc_results} reports the results for $\log_2|\mathcal{V}|^{2}$ number of samples in the case of Rand/Uni-Sample-SM. For Rand/Unigram Hierarchical SM, we use a 2-hidden layer tree with 10 classes per child node. 

\textbf{Baselines for ECOC-NSP Mixture Sampling}
To test Latent Variable Mixture Sampling (LVMS), we directly compare its application in HS and ECOC, two closely related latent NSP methods.
Additionally, we compare the performance of LVMS against the most related sampling-based supervised learning technique called scheduled sampling (SS)~\citep{bengio2015scheduled}. For SS with cross-entropy based training (SS-CE), we also consider using a baseline of the soft-argmax (Soft-SS-CE) where a weighted average embedding is generated proportional to the predicted probability distribution.

\textbf{Evaluation Details} To compute perplexities for ECOC-NSP, the codewords are viewed in terms of a product of marginal probabilities. Hence, when approximating the posterior, we treat latent codes as a factorial distribution~\cite{marlow1965factorial} where each bit is independent of one another ($c_i \perp c_j \; \forall i, j$) . 
Among the $k_i$ error checks corresponding to a particular tokens codeword $C_{k_i}$, we choose the most probable of these checks, as shown in \autoref{eq:top_code}, when computing the binary cross-entropy loss at training time.

\begin{equation}\label{eq:top_code}
  p_{\hat{C}} = \max_{k_i}\prod_{c_{k_i}}^{C_{k_i}}\Big( \phi(x_t, h_t) \Big) , \; i = [0,1..,k]
\end{equation}

At test time, if the predicted codeword $\hat{C}$ falls within the $k$ error-checking bits of codeword $C$, it is deemed a correct prediction and assigned the highest probability of all $k$ predictions. Note that we only convert the ECOC predictions to perplexities to be comparable against baselines (we could use Hamming Distance or Mean Reciprocal Rank when the codes are easily decoded by ordered semantically or by Hamming distance).

\begin{figure}
\centering
  \centering
  \includegraphics[width=.8\linewidth]{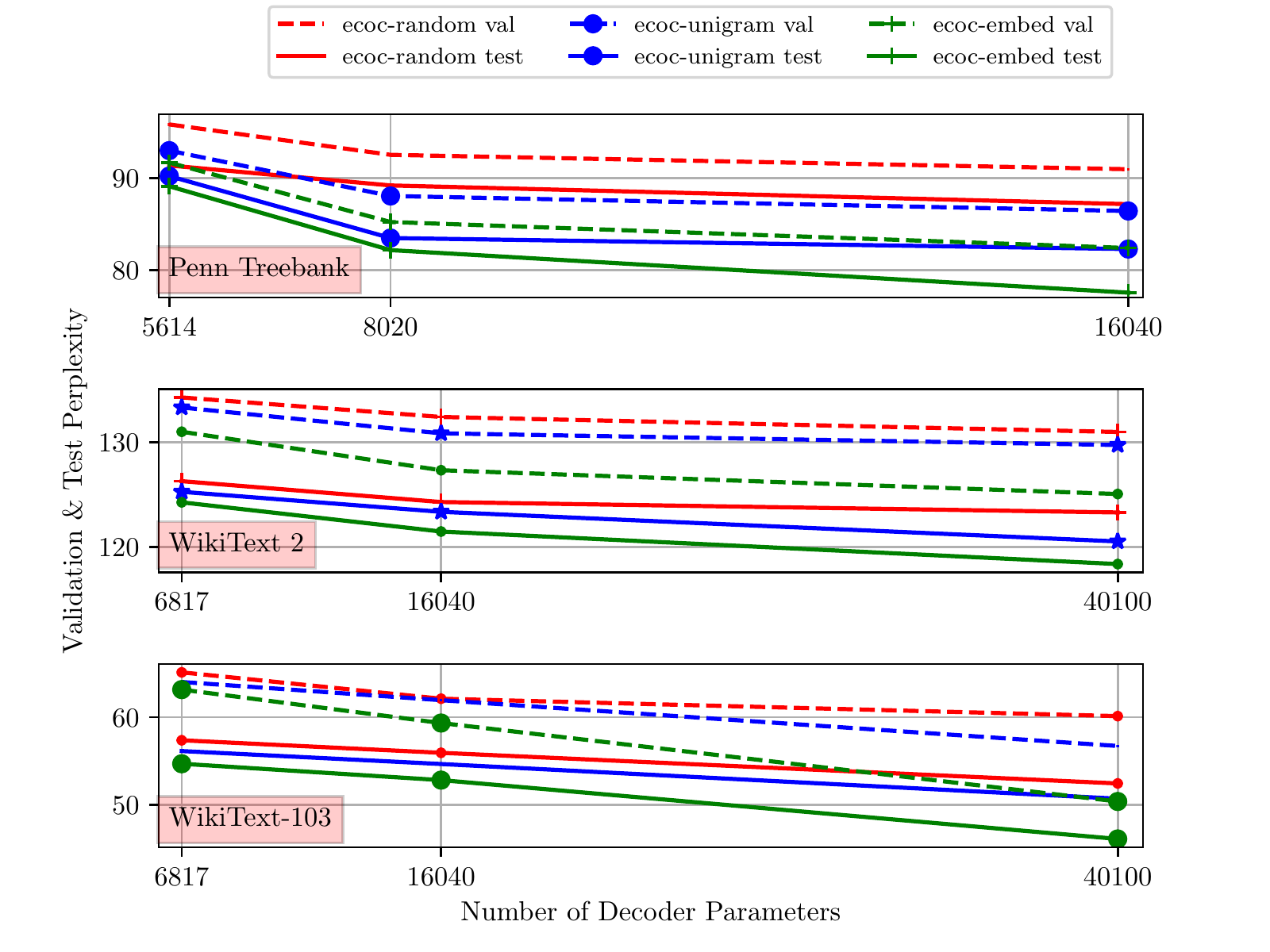}
  \captionof{figure}{\small{ECOC-NSP Peplexity vs. Decoder Parameters (corresponding to 14/20/40 codeword bits for Penn-TreeBank and 17/40/100 codeword bits for WikiText-2/103)}}\label{fig:ppl_decoder_params}
\end{figure}

\section{Results}

\paragraph{Error-Correcting Output Coded NSP}

We first compare our proposed ECOC-NSP to aforementioned methods that approximate softmax normalization, using binary trees and latent codes that are ordered according to unigram frequency (Uni-Hierarchical-SM and Uni-ECOC). This is also the same ordering we use to compare our proposed CMS-ECOC sampling method to scheduled sampling~\citep{bengio2015scheduled} in standard cross-entropy training with softmax normalization. 
Although, these are not directly comaprable, since ECOC-NSP introduces a whole new paradigm, we use the common evaluation measures of Hamming distance and accuracy to have some kind of baseline with which we can compare our proposed method to.
Equation \ref{fig:ppl_decoder_params} shows how the reduction in perplexity as the number of ECOC-LSTM decoder parameters increase as more bits are added to the codeword. For PTB, large perplexity reductions are made between 14-100 codebits, while between 100-1000 codebits there is a gradual decrease. In contrast, we see that there is more gained from increasing codeword size for WikiText-2 and WikiText-103 (which preserve the words that fall within the long-tail of the unigram distribution). We find the discrepancy in performance between randomly assigned codebooks and ordered codebooks is more apparent for large compression ($|C| < |\mathcal{V}|/10$). Intuitively, the general problem of well-separated codes is alleviated as more bits are added.

Equation \ref{tab:ecoc_results} shows that overall ECOC with a rank ordered embedding similarity $\mathcal{C}$ (Embedding-ECOC) almost performs as well as the full-softmax (8.02M parameters) while only using 1000 bits for PTB ($|\mathcal{V}|/20$ and ) and 5K bits for WikiText-2 ($|\mathcal{V}|/25$) and WikiText-103 ($|\mathcal{V}|/30$). 
The HS-based models use a 2-hidden layer tree with 10 tokens per class, resulting in 4.4M parameters for PTB, 22.05M parameters for WikiText-2 (full softmax - 40.1M) and WikiText-103. Moreover, we find there is a consistent improvement in using Embedding-ECOC over using a random codebook (Random-ECOC) and a slight improvement over using a unigram ordered codebook (Uni-ECOC). Note that in both Embedding-ECOC and Uni-ECOC,  the number of error-checking bits are assigned inversely proportional to the rank position when ordering embedding similarities (as discussed in Equation \ref{sec:esbc}) and unigram frequency respectively. We also found that too many bits e.g $|C| = |\mathcal{V}|$ takes much longer ($\epsilon \in$ [20-30] more for PTB) to converge with negligible perplexity reductions. Hence, the main advantage of ECOC-NLVMS is the large compression rate while maintaining performance (e.g PTB with $|C|=40$, there is less than 2 perplexity points compared to the full softmax).

\begin{table}[ht]
\resizebox{1.\linewidth}{!}{%
\begin{tabular}{c|cc|cc|cc}
\toprule[0.2em]
\textbf{Model} & \multicolumn{2}{c}{\textbf{PTB}} &  \multicolumn{2}{c}{\textbf{WikiText-2}} &  \multicolumn{2}{c}{\textbf{WikiText-103}} \\

& Val. & Test & Val. & Test & Val. & Test\\

\midrule

Full SM & \textbf{\emph{86.19}} & \textbf{\emph{79.24}} & \textbf{\emph{124.01}} & \textbf{\emph{119.30}} & \textbf{\emph{56.72}} & \textbf{\emph{49.35}} \\
Rand-Sample-SM & 92.14 & 81.82 & 136.47 & 129.29 & 68.95 & 59.34 \\
Uni-Sample-SM & 90.37 & 81.36 & 133.08 & 127.19 & 66.23 & 57.09 \\
Rand-Hierarchical-SM & 94.31 & 88.50 & 133.69 & 127.12 & 62.29 & 54.28 \\
Uni-Hierarchical-SM & 92.38 & 86.70 & 130.26 & 124.83 & 62.02 & 54.11 \\
Adaptive-SM & 91.38 & 85.29 & 118.89 & 120.92 & 60.27 & 52.63 \\
NCE & 96.79 & 89.30 & 131.20 & 126.82 & 61.11 & 54.52 \\

\midrule[0.15em]
Random-ECOC & 91.00 & 87.19 & 131.01 & 123.29 & 56.12 & 52.43 \\
Uni-ECOC & 86.44 & 82.29 & 129.76 & 120.51 & \textbf{\emph{52.71}} & \textbf{\emph{48.37}} \\
Embedding-ECOC & \textbf{\emph{84.40}} & \textbf{\emph{77.53}} & \textbf{\emph{125.06}} & \textbf{\emph{120.34}} & 57.37 & 49.09 \\

\bottomrule[0.2em]
\end{tabular}%
}
   \caption{\small{LSTM Perplexities for Full Softmax (SM), Sample-Based SM (Sample-SM), Hierarchical-SM (HSM), Adaptive-SM, NCE and ECOC-NSP.}
   \iffalse Accuracy and Hamming Distance\fi}
  \label{tab:ecoc_results}
\end{table}

\paragraph{Latent Variable Mixture Sampling Text Generation Results}

\begin{table}
\resizebox{1.\linewidth}{!}{%
\begin{tabular}{c|cc|cc|cc}
\toprule[0.2em]
 & \multicolumn{1}{c}{\textbf{B1}} &  \multicolumn{1}{c}{\textbf{B2}} &  \multicolumn{1}{c}{\textbf{B3}} & \multicolumn{1}{c}{\textbf{B4}} &  \multicolumn{1}{c}{\textbf{R-L}} &  \multicolumn{1}{c}{\textbf{MET}} \\

\midrule
Full-SM & 71.09 & 51.33 & 32.85 & 24.67 & 50.28 & 52.70 \\
SS-SM & 73.23 & 52.81 & 33.37 & 26.11 & 52.60 & 54.51 \\
Soft-SS-SM & \textbf{\emph{73.54}} & \textbf{\emph{53.01}} & \textbf{\emph{33.26}} & \textbf{\emph{27.13}} & \textbf{\emph{54.49}} & \textbf{\emph{54.83}} \\
\midrule
SS-Adaptive-SM & 70.45 & 50.22 & 31.38 & 23.59 & 51.88 & 51.83 \\
\midrule
SS-Hierarchical-SM & 67.89 & 48.42 & 30.37 & 22.91 & 49.39 & 50.48 \\
CLVMS-Hierarchical-SM & 69.70 & 49.52 & 31.91 & 24.19 & 51.35 & 51.20 \\
DLVMS-Hierarchical-SM & 71.04 & 50.61 & 32.26 & 24.72 & 52.83 & 52.36 \\
\midrule
SS-ECOC & 72.02 & 52.03 & 32.57 & 25.42 & 51.39 & 53.51 \\
Soft-SS-ECOC & 72.78 & 53.29 & 33.15 & 25.93 & 52.07 & 54.22 \\
\midrule[0.15em]

CLVMS-ECOC & \textbf{\emph{74.70}} & \textbf{\emph{53.09}} & \textbf{\emph{34.28}} & \textbf{\emph{27.05}} & \textbf{\emph{53.67}} & \textbf{\emph{55.62}} \\

DLVMS-ECOC & \textbf{\emph{74.92}} & \textbf{\emph{53.56}} & \textbf{\emph{34.70}} & \textbf{\emph{27.81}} & \textbf{\emph{54.02}} & \textbf{\emph{55.85}} \\

\bottomrule[0.2em]
\end{tabular}%
}
   \caption{MSCOCO Test Results on BLEU (B), ROUGE-L (R-L) \& METEOR (MET) Eval. Metrics}\label{tab:coco_results} 
\end{table}


Table \ref{tab:coco_results} shows all results of LVMS when used in HS and ECOC-based NSP models for the MSCOCO image captioning dataset~\cite{lin2014microsoft} using the Karpathy validation and test splits using a beam search with a width of 5 at test time. We use $|c|=200$ to account for vocabulary size $|V|=10^3$, leaving $|c|-\log_2(|V|)=186$ error-check bits leftover $\forall C \in \mathcal{C}$. The HS uses the Categorical Concrete distribution for DLVMS-HS and Binary Concrete Distribution for DCMS-ECOC. In our experiments we found $\tau=2.0$ to be the upper threshold from an initial grid search of $\tau \in \{0.2, 0.5, 1, 2, 5, 10\}$ for both DLVMS-HS and DCMS-ECOC, where $\tau < 2$ corresponds to little exploration and $\tau > 2$ results in too much exploration, particularly early on when the model is performing larger gradient updates. CLVMS-Hierarchical-SM and CLVMS-ECOC both monotonically increases $\tau$ according to Equation \ref{eq:cur_exp}. 

Both HS and ECOC use Embedding ordered decoder matrix (we omit the -Embedding extension).
We baseline this against both SS and the soft-argmax version of SS, the most related sample-based supervised learning approach to LVMS. Furthermore, we report results on CLVMS-ECOC (Curriculum-LVMS ECOC) which mixes between true targets and codewords predictions according to the schedule in Equation \ref{eq:cur_exp} and a differentiable extension of LVMS via samples from the Gumbel-Softmax (DCMS-ECOC). For both DCMS-ECOC and DLVMS-Hierarchical-SM we sample from each softmax defined along the path to the target code at training time.
We find that using a curriculum in CLVMS-ECOC with a semantically ordered codebook outperforms the full softmax that uses scheduled sampling (SS-SM) and its weighted-variant (Soft-SS-SM).
Moreover, DLVMS-ECOC further improves over CLVMS-ECOC on MSCOCO. LVMS makes a consistent improvement over SS. This suggests that LVMS is an effective alternative for latent variable based NSPs in particular (mixture sampling is ill-suited to one-hot targets as they are extremely sparse).

\section{Conclusion}
This work proposed an error-correcting neural language model and a novel Latent Variable Mixture Sampling method for latent variable models. 
We find that performance is maintained compared to using the full conditional and related approximate methods, given a sufficient codeword size to account for correlations among classes. This corresponds to 40 bits for PTB and 100 bits for WikiText-2 and WikiText-103. Furthermore, we find that performance is improved when rank ordering the codebook via embedding similarity where the query is the embedding of the most frequent word.
\newline
Lastly, we introduced Latent Variable Mixture Sampling to mitigate exposure bias. This can be easily integrated into training latent variable-based language models, such as the ECOC-based language model. We show that this method outperforms the well-known scheduled sampling method with a full softmax, hierarchical softmax and an adaptive softmax on an image captioning task, with less decoder parameters than the full softmax with only 200 bits, 2\% of the original number of output dimensions.


{\small \bibliography{emnlp2019}|\small \bibliographystyle{acl_natbib}|}

\end{document}